\documentclass[10pt,twocolumn,letterpaper]{article}

\usepackage{cvpr}              
\definecolor{cvprblue}{rgb}{0.21,0.49,0.74}
\usepackage[pagebackref,breaklinks,colorlinks,allcolors=cvprblue]{hyperref}

\usepackage{amsmath}
\usepackage{amssymb}
\usepackage{algorithm}
\usepackage{algorithmic}
\usepackage{multirow}

\def\paperID{32625} 
\def\confName{CVPR}
\def\confYear{2026}

\title{Ego-InBetween: Generating Object State Transitions in Ego-Centric Videos
}

\author{                                        
Mengmeng Ge$^{1,}$\thanks{Equal contribution.} \qquad Takashi Isobe$^{1,}$\footnotemark[1] \thanks{Corresponding author.} \qquad
Xu Jia$^{2,}$\footnotemark[2] \qquad
Yanan Sun$^{3}$ \qquad 
Zetong Yang$^{4}$ \\ 
Weinong Wang$^{5}$ \qquad
Dong Zhou$^{1}$ \qquad
Dong Li$^{1}$ \qquad
Huchuan Lu$^{2}$ \qquad
Emad Barsoum$^{1}$ \\
$^{1}$ Advanced Micro Devices, Inc. \qquad $^{2}$ Dalian University of Technology \\
$^{3}$ The Hong Kong University of Science and Technology \qquad $^{4}$ The Chinese University of Hong Kong \\
$^{5}$ Xi'an Jiaotong University \\
{\tt\small \{k15201363625,panisobe,jiayushenyang\}@gmail.com}
}

\begin{document}

\maketitle

\begin{abstract}

Understanding physical transformation processes is crucial for both human cognition and artificial intelligence systems, particularly from an egocentric perspective, which serves as a key bridge between humans and machines in action modeling. We define this modeling process as Egocentric Instructed Visual State Transition (EIVST), which involves generating intermediate frames that depict object transformations between initial and target states under a brief action instruction. EIVST poses two challenges for current generative models: (1) understanding the visual scenes of the initial and target states and reasoning about transformation steps from an egocentric view, and (2) generating a consistent intermediate transition that follows the given instruction while preserving object appearance across the two visual states. To address these challenges, we propose the EgoIn framework. It first infers the multi-step transition process between two given states using TransitionVLM, fine-tuned on our curated dataset to better adapt to this task and reduce hallucinated information. It then generates a sequence of frames based on transition conditions produced by the proposed Transition Conditioning module. Additionally, we introduce Object-aware Auxiliary Supervision to preserve consistent object appearance throughout the transition. Extensive experiments on human-object and robot-object interaction datasets demonstrate EgoIn's superior performance in generating semantically meaningful and visually coherent transformation sequences.

\end{abstract}    
\section{Introduction}

\begin{figure*}[ht]
	\centering
	\includegraphics[width=1\textwidth]{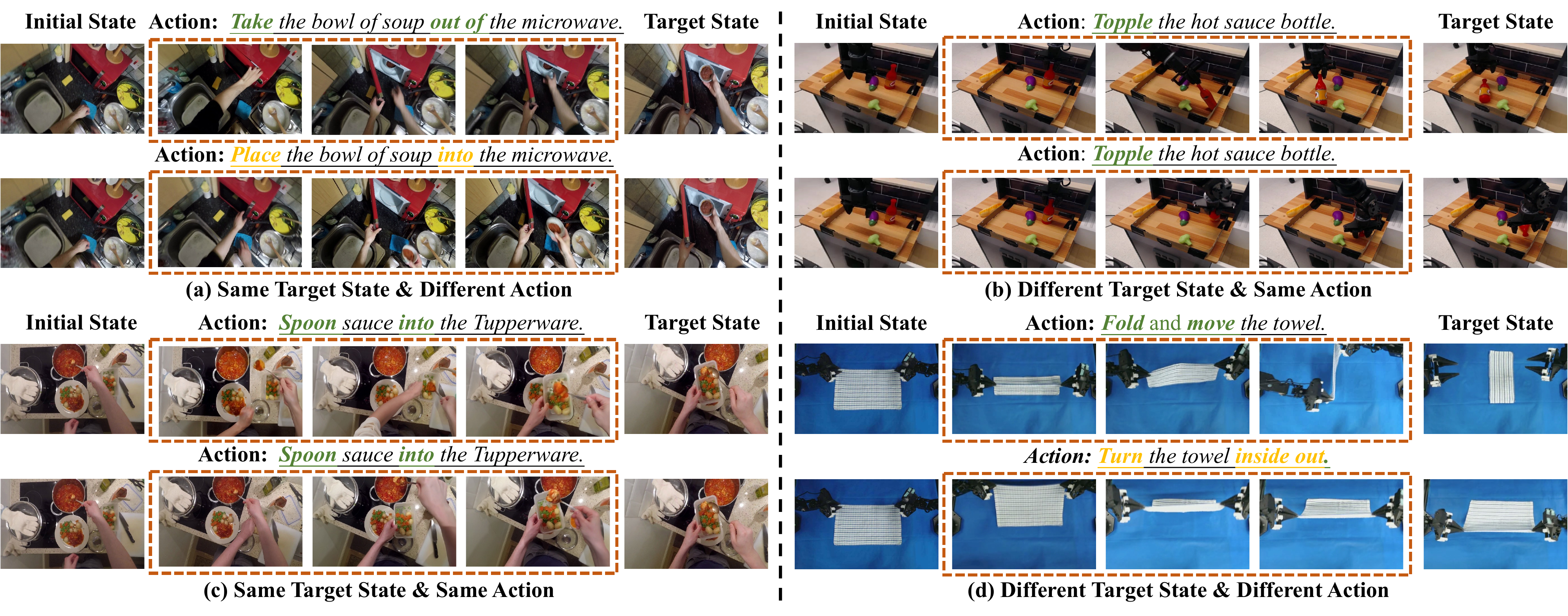}
        \caption{Examples of diverse generated object state transition sequences under different textual and visual conditions: (a) different action instructions with the same initial and target states; (b) different target states with the same initial state and action instruction; (c) the same action instruction, initial state, and target state; (d) different action instructions and target states with the same initial state.}
	\label{fig:motivation}
\end{figure*}

As humans, our understanding of the physical world is deeply rooted in our ability to perceive how states evolve over time. We do not merely observe objects in isolation but intuitively reason about the processes that connect them, forming mental models of cause and effect. This innate curiosity about how things change shapes how we learn and teach. The same capability is essential for visual assistants and robots, which must grasp transformation processes to effectively assist humans.
While recent advances in Vision-Language Models (VLMs) have enabled the generation of abstract textual descriptions or instructions for transformation processes, these instructions often lack the detail necessary to fully describe complex physical transformations. This limitation highlights the need for ``visual instruction'' to provide visual guidance for transformation processes by generating intermediate frames, bridging the gap between high-level concepts and detailed execution.

In this paper, we seek to bridge the gap between human cognition and machine understanding of state transformations from an egocentric perspective by introducing the Egocentric Instructed Visual State Transition (EIVST) task. The task requires a generative model to produce a coherent sequence of intermediate frames that capture an object's transformation from an initial state to a target state, guided by an action instruction.
EIVST has potential applications in areas such as robotic manipulation, computer-aided design, and interactive educational tools, where the ability to visualize complex object transformations is crucial. 

EIVST shares certain similarities with 
the task of text-conditioned video prediction (TVP) where future video frames are predicted based on a language instruction and a reference frame. 
Seer~\cite{gu2023seer} and AID~\cite{xing2024aid} focus on predicting future frames to visualize the action exerted on the object and the outcome of that action. However, most of them condition video generation on a single reference frame and an instruction, without visual guidance from the target state to indicate where the generation process should lead.
In addition, it relates to tasks of transition video generation from images, such as video frame interpolation~\cite{reda2022film,danier2024ldmvfi,jain2024video,shen2024dreammover} and scene transition~\cite{chen2023seine,zeng2024make,zhang2024diffmorpher,zhang2024tvg}, with the purpose of generating smooth motion or appearance transitions between two reference images. 
They do not focus on action-induced transformations and therefore may fail to capture semantically complex changes in the objects. Different from these tasks, EIVST requires semantic understanding of visual scenes and reasoning capability over the transformation process between given frames, as shown in Fig.~\ref{fig:motivation}. 

We propose the EgoIn framework to address the challenges posed by EIVST. EgoIn follows a divide-and-conquer strategy, which first infers the multi-step transition process between two visual anchors, and then generates a sequence of frames based on the transition conditions. 
Specifically, we adapt a VLM for transition process modeling by tuning it on data curated by prompting GPT-4o with specially designed inputs (e.g., object bounding boxes and entire video frames) and corresponding instructions. The model is fine-tuned to describe both the initial and target object states while also inferring K intermediate steps and their corresponding temporal ranges. 
Subsequently, we introduce the Transition Conditioning (TC) module, which integrates multimodal state-aware features and inferred transition-aware features as frame-wise conditions to guide transition process generation. These conditions are then injected into the denoising U-Net via a frame-wise weighted cross-attention mechanism, ensuring transition-aware controllability in video generation. 
To further maintain consistent object appearance and smooth motion during state transitions in a video, we propose the Object-Aware Auxiliary Supervision (OAS) as an additional loss. It is designed to localize key operated objects in the process and work together with video reconstruction loss to promote model learning. Extensive experiments are conducted on several egocentric human-object interaction and object manipulation datasets. Qualitative and quantitative evaluations show that the proposed EgoIn achieves superior performance on the EIVST task, surpassing existing video generative models even after they are finetuned on our dataset.
The main contributions are summarized as follows:
\begin{itemize}
    \item We introduce the Egocentric Instructed Visual State Transition (EIVST) task, in which generative models are required to produce a coherent sequence of frames that depict the transformation of an object’s state from the initial to target states, conditioned on the action instruction.
    \item We propose the EgoIn framework to address the challenges posed by EIVST, integrating three key innovations:
    (1) TransitionVLM, which interprets initial state, target state, and action instruction to infer the transition steps between visual anchors through domain-specific fine-tuning;
    (2) Transition Conditioning (TC), which generates frame-wise transition conditions to guide the video generation process; and
    (3) Object-aware Auxiliary Supervision (OAS), which further enhances visual fidelity and object consistency in the synthesized transitions.
    \item Extensive experiments on egocentric human–object and robot–object interaction datasets demonstrate the effectiveness of the proposed EgoIn framework.
\end{itemize}

\section{Related Work}
\begin{figure*}[t]
	\centering
	\includegraphics[width=1\textwidth]{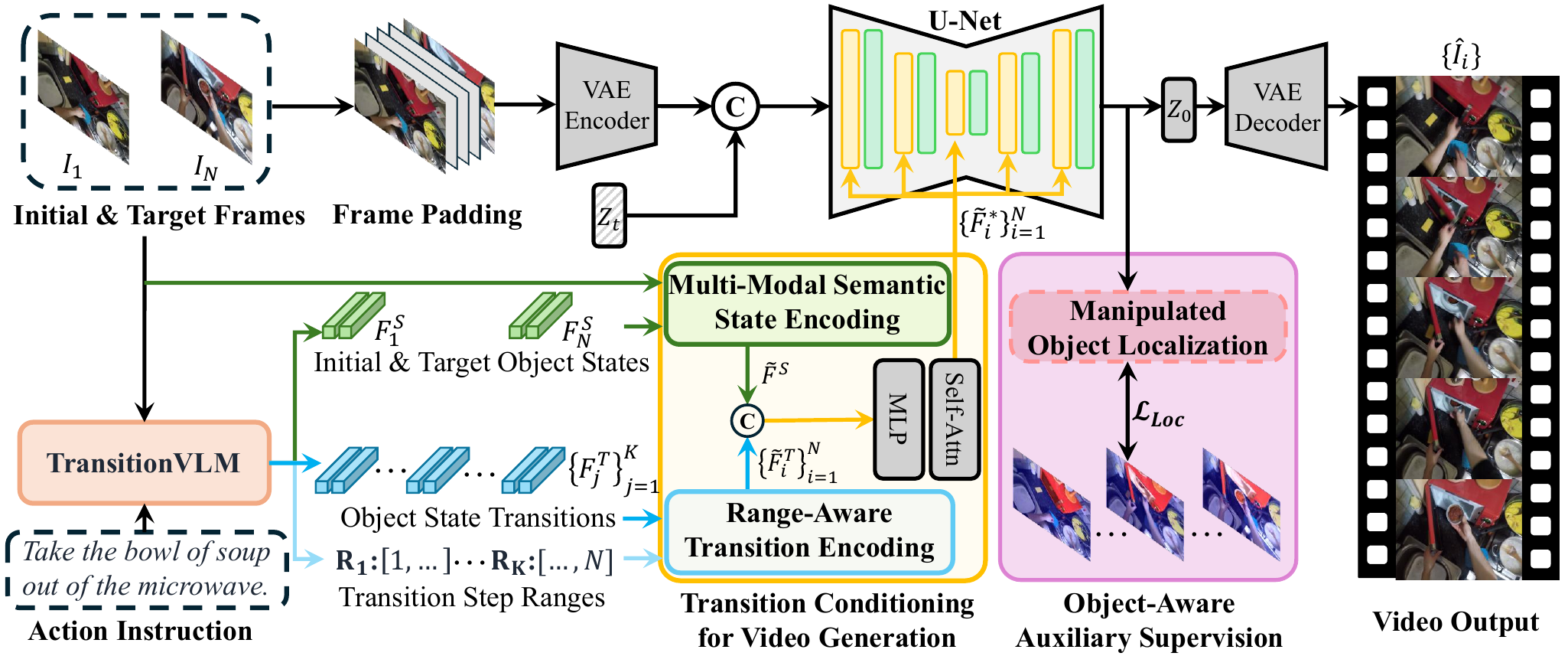}
	\caption{Illustration of the proposed EgoIn framework. EgoIn works in two stages: (1) transition process modeling using the tuned TransitionVLM and (2) intermediate frame generation based on the transition conditions. Additionally, Object-aware Auxiliary Supervision is designed to localize key objects involved in actions and works with the frame reconstruction loss to promote model learning.}
	\label{fig:framework}
\end{figure*}

\noindent\textbf{Conditional Video Generation.}
Video generation models have advanced rapidly in recent years~\cite{he2022latent,wang2023lavie,blattmann2023svd,zhang2023i2vgen,isobe2025amd,xing2025dynamicrafter}. Building on these pioneering works, recent methods focus on more specialized tasks. GenHowto~\cite{souvcek2024genhowto} and ShowHowto~\cite{souvcek2025showhowto} generate discrete key frames of an object's future states in a step-by-step manner, but their results lack temporal coherence, which is essential for aligning with human understanding of physical processes. A few works~\cite{gu2023seer,xing2024aid} attempt to predict coherent future frames from a single state by extending the capabilities of I2V models. However, these methods struggle to reach the desired target state. For example, if the given frame shows a closed refrigerator, it is difficult to infer its contents when attempting to take a bowl of lettuce from it. This task requires I2V models to infer intermediate transition steps and generate frames guided by those inferred results. In addition, the generated frames should follow the given actions and remain consistent with the two visual anchor frames. This more controllable setting makes the task suitable for studying the underlying principles of how the physical world operates and the demonstration capabilities required in embodied AI.

\noindent\textbf{Transition Video Generation.} Existing methods for intermediate frame generation can be categorized into two main directions. One direction is video transition modeling~\cite{chen2023seine,zhang2024diffmorpher, zhang2024tvg, zeng2024make, wang2024generative, feng2025explorative, kan2025sage, bi2025mobius}, which aims to model smooth appearance changes between the initial and target frames. However, they struggle to generate complex motion patterns between the given states when required. 
The second direction is video frame interpolation (VFI)~\cite{voleti2022mcvd, wang2023interpolating, danier2024ldmvfi}. The rapid advancements in diffusion models have led to significant breakthroughs in handling large motion between two frames~\cite{wang2024generative,feng2025explorative,yang2025vibidsampler,zhang2025motion,guo2025controllable, zhang2025egvd, zhu2025generative}. However, these models often produce simple or repetitive motion patterns, and struggle to generate sequences involving multi‑step and compositional actions. This limitation mainly arises from being overly reliant on additional guidance conditions~\cite{guo2025controllable,zhu2025generative} and the lack of reasoning capability to infer the potential transition process between two object states. 

\noindent\textbf{VLM in Generative Diffusion Models.}
Vision-Language Models (VLMs)~\cite{liu2023llava, achiam2023gpt, chen2024sharegpt4video, bai2025qwen2, li2026seek} possess strong reasoning capabilities for processing and understanding images. Recently, this capability has been introduced for image generation~\cite{wu2023self,lian2023llmgrounded,lai2024lego} and video generation~\cite{lin2023videodirectorgpt,huang2024free, li2024vstar, xing2024aid, xiang2024pandora}. Free-Bloom~\cite{huang2024free} and VSTAR~\cite{li2024vstar} utilize VLMs to decompose video prompts into detailed frame-by-frame descriptions, while AID~\cite{xing2024aid} employs a VLM to predict future action steps. However, directly applying VLMs to specific tasks without tuning is prone to generating plausible-sounding answers without factual basis. To address this issue, we fine-tune the VLM on curated data, ensuring reliable state transition modeling and further improving the performance of the I2V model.

\section{Method}
\label{sec:method}
\subsection{Overall} 
\begin{figure*}[t]
	\centering
	\includegraphics[width=1\textwidth]{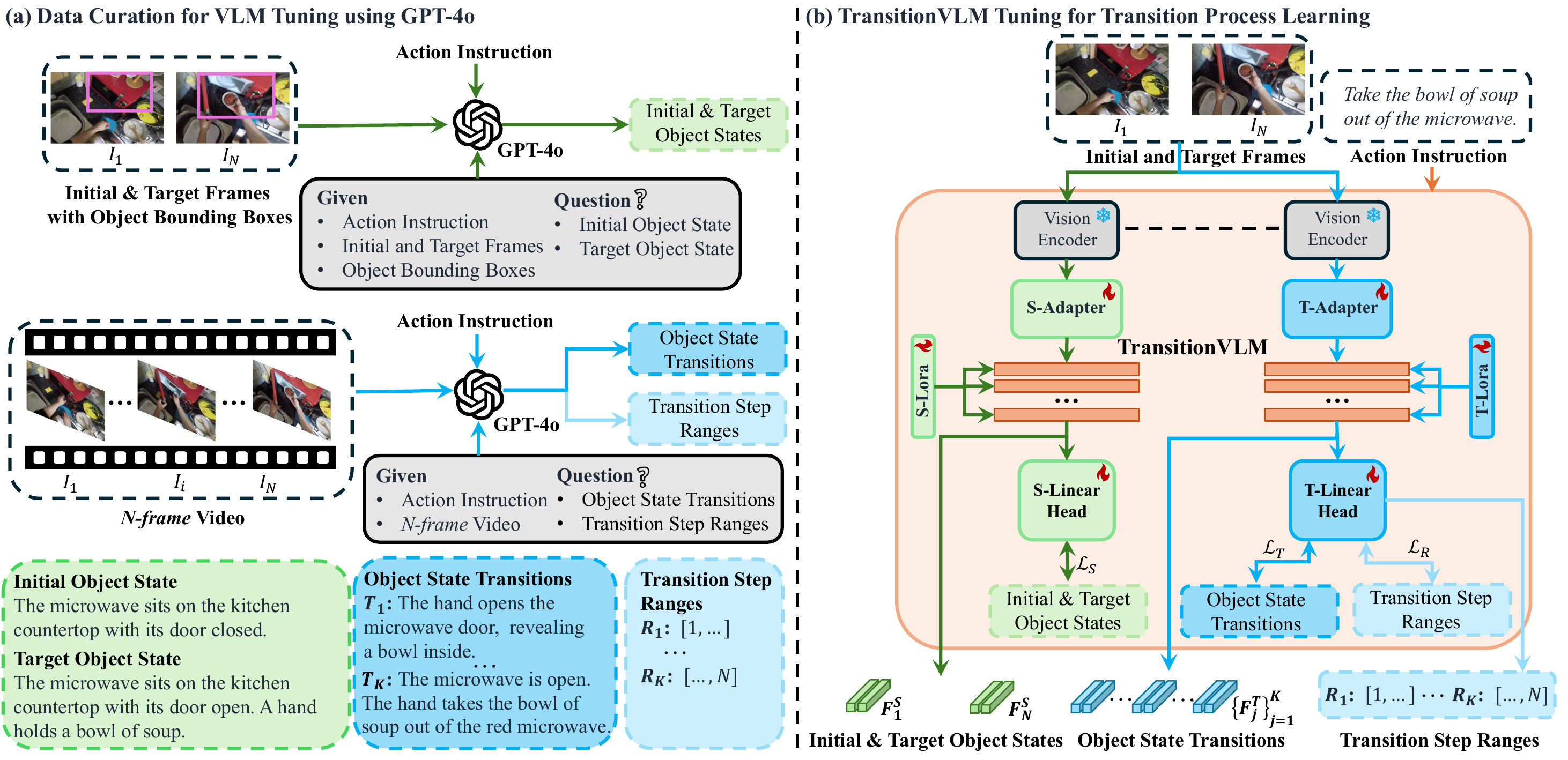}
	\caption{Illustration of the tuning process for the proposed TransitionVLM: (a) shows the data curation process used to obtain state-aware and transition-aware information by providing a specific input to GPT-4o. (b) illustrates the VLM tuning process using the curated data.}
	\label{fig:llm}
\end{figure*}
We first formally define the Egocentric Instructed Visual State Transition (EIVST) task. Given an initial frame $I_1$ (initial state), a target frame $I_N$ (target state), and a brief textual instruction describing the applied action, the goal is to generate a sequence of intermediate frames $\{\hat{I}_i \mid 2 \leq i \leq N-1\}$ that illustrate the state transition process over $K$ inferred steps. To address this challenge, we propose the EgoIn framework, as illustrated in Fig.~\ref{fig:framework}. It first infers the multi-step transition process between the initial and target states via the tuned TransitionVLM. The produced transition conditions are then integrated via the proposed Transition Conditioning (TC) module and injected into the U-Net’s cross-attention layers in a frame-wise manner, guiding the generation of transition frames. In addition, we apply Object‑aware Auxiliary Supervision (OAS) during training to ensure consistent object appearance in manipulated objects throughout the transition.

\subsection{VLM for Transition Process Modeling}
\label{sec:3.3}
Existing Image-to-Video (I2V) generative models tend to overly rely on the provided information, limiting their ability to further explore and reason beyond the given cues. In contrast, humans leverage a broader range of contextual information, enabling them to reason about which state transitions occur, what transformations take place, and where they happen between two object states. This motivates us to introduce a VLM to guide I2V models in addressing this limitation. Inspired by how humans reason about object transitions, we leverage a VLM to reason about three key aspects of a state transition: (1) \textbf{which} objects are being manipulated, inferred from appearance changes between the initial and target frames and the action instruction;
(2) \textbf{what} intermediate transition steps occur between these states; and
(3) \textbf{where} these transition steps should be temporally grounded within the sequence.

\noindent\textbf{Data Curation for VLM Tuning.} Directly using VLMs to predict these results presents several challenges. While trained on large-scale datasets and equipped with strong reasoning capabilities, they are not explicitly designed for transition modeling. Consequently, they are prone to generating plausible-sounding textual outputs that are not grounded in the input images, extending transition steps beyond the target frame and uniformly allocating temporal ranges across all transitions.
To address these challenges, we fine-tune the Vision-Language Model (VLM) using curated data from GPT-4o~\cite{achiam2023gpt}. We design two types of input formats by prompting GPT-4o with specially designed instructions for states and transitions, as shown in Fig.~\ref{fig:llm}(a). For state-aware descriptions, we incorporate bounding boxes to guide GPT-4o in focusing on specific object regions, thereby reducing irrelevant background details. These bounding boxes are generated by first identifying the manipulated objects from the action instruction and then detecting them using Qwen2.5-VL~\cite{bai2025qwen2}. For transition-aware descriptions, we provide GPT-4o with the entire video sequence to generate a set of $K$ state transitions and the corresponding temporal ranges.

It is worth noting that directly prompting GPT-4o with only the initial and target frames often leads to redundant, vague, or task-irrelevant descriptions, as the model tends to hallucinate imaginary transitions that are not visually grounded. To obtain concise and visually reliable transition descriptions, \textbf{we adopt two prompting strategies based on auxiliary information available only at data generation}:
(1) supplying bounding boxes of key objects, obtained via open-vocabulary detection using Qwen2.5-VL guided by the action instruction, to produce state-aware descriptions; and
(2) providing the full $N$-frame video sequence, enabling GPT-4o to generate transition-aware descriptions that more accurately capture intermediate state changes than using only the initial and target frames.

\noindent\textbf{VLM Tuning for Transition Process Learning.}
In this paper, we use Qwen2.5-VL 7B~\cite{bai2025qwen2} as the base VLM due to its efficiency and low computational cost. Fig.~\ref{fig:llm}(b) illustrates the tuning process of the proposed TransitionVLM. The key insight of this fine-tuning process is to reduce ungrounded or fabricated information while preserving the reasoning and generalization capability of the original model. To achieve this, we fine-tune the VLM using LoRA instead of full fine-tuning. We inject an S-Adapter, S-LoRA, and a prediction head into the original model and fine-tune these components with curated data by GPT-4o to learn state-aware information (i.e., the initial and target object states). Similarly, we introduce a T-Adapter, T-LoRA, and another prediction head to learn transition-aware information (i.e., $K$-step object state transitions and their corresponding temporal ranges). The tuned TransitionVLM substantially reduces fabricated information, as shown in Fig.~\ref{fig:VLM-case}.

\begin{figure}[t]
	\centering
	\includegraphics[width=0.48\textwidth]{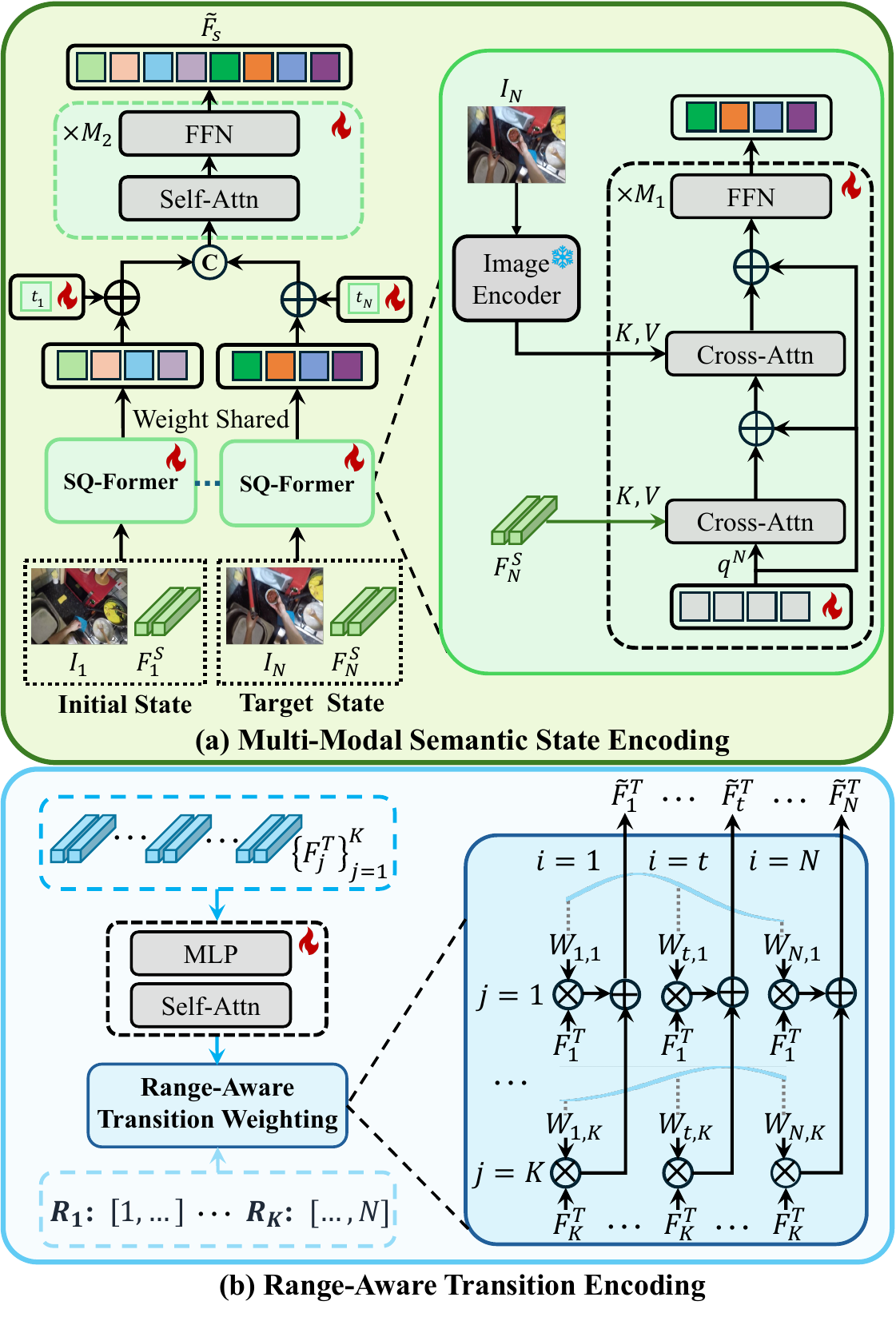}
 		\vspace{-0.2in}
	\caption{Illustration of the Transition Conditioning (TC) module, which consists of two components:
(a) Multi-modal Semantic State Encoding, which extracts state-aware features grounded in visual semantics; and
(b) Range-Aware Transition Encoding, which generates and refines transition-aware features for each frame based on the inferred step ranges.}
	\label{fig:qformer}
 		\vspace{-0.2in}
\end{figure}

\subsection{Transition Conditioned Video Generation}
Existing I2V models depend on attention layers to implicitly model the input conditions, and the lack of explicit frame-wise guidance makes it difficult to determine proper temporal ranges for transitions, often causing certain steps to be missed. To address this limitation, we adapt the outputs of TransitionVLM to the I2V model through a frame-wise cross-attention mechanism. The key idea is to explicitly incorporate frame-wise transition conditions to enable transition-aware video generation while preserving the semantic content of the initial and target frames. To this end, we introduce the Transition Conditioning (TC) module, which extracts state-aware transition-aware features that capture the underlying state transformation process.

\noindent\textbf{Multi-modal Semantic State Encoding.}
The state-aware features are extracted as semantic representations, as illustrated in Fig.~\ref{fig:qformer}(a). This module takes two frames, $\{I_1, I_N\}$, along with the outputs of TransitionVLM that describe the initial and target object states. 
Rather than directly using the textual prompts generated by TransitionVLM, we leverage the features extracted before the prediction heads, denoted as $F^S_1$ and $F^S_N$, as they encapsulate richer information and are more robust to noise. 
The input is first divided into the initial state $\{I_1, F^S_1\}$ and the target state $\{I_N, F^S_N\}$. 
These two sets are processed by two weight-shared SQ-Formers, inspired by BLIP~\cite{li2023blip}, to align visual and semantic features and produce state-aware representations. 
To fully leverage their complementary information, these features are then fused through stacked layers of Self-Attention (Self-Attn) and Feed-Forward Networks (FFN), yielding $\widetilde{F}^S$. 
Additionally, we introduce learnable position tokens to effectively distinguish the initial and target state positions and reinforce temporal awareness.

\begin{figure*}[t]
	\centering
	\includegraphics[width=1\textwidth]{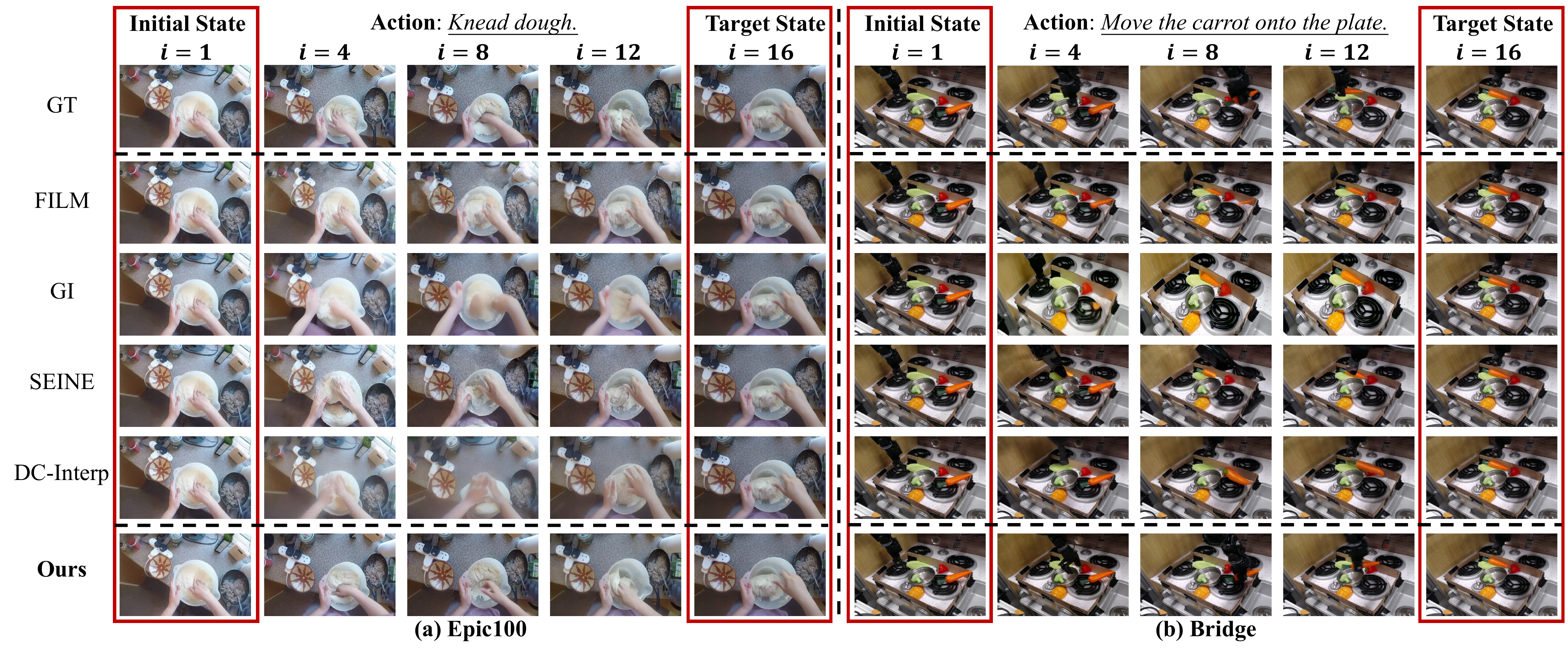}
	\caption{Qualitative comparison on Epic100 and Bridge. Intermediate frames ($i = 4, 8, 12$) from the generated sequences are shown.}
	\label{fig:compare}
	\vspace{-0.2in}
\end{figure*}
\noindent\textbf{Range-Aware Transition Encoding.} 
We further generate and refine frame-wise transition features for the I2V model, as shown in Fig.~\ref{fig:qformer}(b). This refinement is essential for capturing controllable, transition-aware signals in video generation. We introduce a softened weighting mechanism that adaptively reweights transitions by resampling weights from a Gaussian distribution based on the predicted step ranges, smoothing noisy boundaries and producing more stable transition representations. Based on the transition step ranges $\{R_j\}_{j=1}^K$, we resample weights $W_{i,j}$ from a Gaussian distribution for each frame $i$ and step $j$, where the center and length of $R_j$ serve as the mean and standard deviation. For each frame, the transition features $\{F^T_1,\dots,F^T_K\}$ are weighted by $\{W_{i,1},\dots,W_{i,K}\}$ and summed to form the refined transition condition. We then concatenate $\widetilde{F}^S$ with $\{\widetilde{F}^T_i\}_{i=1}^N$ and process them through an MLP and a Self-Attn layer to obtain the final transition conditions $\{\widetilde{F}^*_i\}_{i=1}^N$. These conditions are injected into the denoising U\text{-}Net via frame-wise cross-attention for $N$-frame video generation.

\subsection{Object-Aware Auxiliary Supervision}
\label{sec:3.4}

To enhance the appearance consistency and motion smoothness of objects across transition steps in the generated video, we propose a multi-task learning strategy that enables the I2V model to learn both frame reconstruction and key object localization through auxiliary supervision. Specifically, we introduce a localization head, consisting of two stacked convolutional layers placed after the last convolutional block of the U-Net. We obtain ground-truth object localization by first using Qwen2.5-VL to identify objects based on the given action instruction. Then, SAM2~\cite{ravi2024sam} generates object localization masks for each frame, following mask-guided generation methods~\cite{jia2024customizing, yuan2025identity}, and the masks are subsequently downsampled to match the output resolution of the localization head. The auxiliary supervision minimizes the mean pixel-wise cross-entropy loss between the predicted probability maps and the downsampled ground-truth localization masks. Notably, SAM2 is used only for ground-truth generation during training, and we do not use localization masks or bounding boxes during inference avoiding any additional computational overhead. The reconstruction loss is conditioned on the initial and target frames $\{I_1, I_N\}$, the frame rate $\text{fr}$, and the frame-wise transition conditions $\{\widetilde{F}^*_i\}_{i=1}^N$. The overall optimization objective is formulated as:
\begin{equation*}
    \min_{\theta} \left(\mathbb{E}_{z, t, \epsilon}\| \epsilon - \epsilon_{\theta}(z_t;  t, \text{fr}, I_{1}, I_{N}, \{\widetilde{F}^*_i\}_{i=1}^N) \|_2^2 + \lambda {\mathcal{L}_{LOC}} \right),
\end{equation*}

\noindent where $\theta$ represents the model parameters, $z$ denotes latent variables, $t$ is the time-step in the diffusion process, and $\epsilon$ is the noise added in the diffusion process. The hyperparameter $\lambda$ controls the weight of the localization loss and is empirically set to 0.1 to balance the two learning objectives while ensuring training stability.

\section{Experiment}
\begin{table*}[ht]
\centering
\setlength{\tabcolsep}{2pt}
\resizebox{\textwidth}{!}{
\begin{tabular}{c|cccc|cccc|cccc|cccc}
\hline
\multirow{2}{*}{Method} & \multicolumn{4}{c|}{\textbf{Epic100}} & \multicolumn{4}{c|}{\textbf{EgoFHO}} & \multicolumn{4}{c|}{\textbf{DualArm}} & \multicolumn{4}{c}{\textbf{Bridge}} \\
 & FVD$\downarrow$ & VTQ$\uparrow$ & VTC$\uparrow$ & VIC$\uparrow$  & FVD$\downarrow$ & VTQ$\uparrow$ & VTC$\uparrow$ & VIC$\uparrow$ & FVD$\downarrow$ & VTQ$\uparrow$ & VTC$\uparrow$ & VIC$\uparrow$ & FVD$\downarrow$ & VTQ$\uparrow$ & VTC$\uparrow$ & VIC$\uparrow$ \\ \hline
FILM & 581.09 & 0.8453 & 0.1694 & 0.9154 & 630.49 & 0.8391 & 0.1668 & 0.9224 & 702.81 & 0.8467 & 0.1713 & 0.9207 & 765.54 & 0.8663 & 0.1695 & 0.9231 \\
TRF & 412.43 & 0.8660 & 0.1746 & 0.9016 & 418.91 & 0.8653 & 0.1725 & 0.9052 & 455.37 & 0.8729 & 0.1768 & 0.9098 & 410.48 & 0.8967 & 0.1781 & 0.9163 \\
GI & 325.48 & 0.8720 & 0.1808 & 0.9175 & 357.41 & 0.8723 & 0.1783 & 0.9216 & 382.44 & 0.8804 & 0.1826 & 0.9283 & 370.29 & 0.9046 & 0.1836 & 0.9320 \\
\hline
SEINE & 343.96 & 0.8683 & 0.1951 & 0.8844 & 316.63 & 0.8698 & 0.1985 & 0.9170 & 341.29 & 0.8771 & 0.1994 & 0.9215 & 322.40 & 0.9090 & 0.2015 & 0.9231 \\ 
DC-Interp & 296.67 & 0.8797 & 0.2041 & 0.9081 & 290.30& 0.8740 & 0.2079 & 0.9196 & 298.51 & 0.8895 & 0.2107 & 0.9324 & 287.47& 0.9153 & 0.2092 & 0.9302 \\
\hline \textbf{Ours} & \textbf{215.27} & \textbf{0.9081} & \textbf{0.2373} & \textbf{0.9313} & \textbf{203.85} & \textbf{0.8987} & \textbf{0.2340} & \textbf{0.9396} & \textbf{209.63} & \textbf{0.9142} & \textbf{0.2361} & \textbf{0.9468} & \textbf{191.09} & \textbf{0.9374} & \textbf{0.2395} & \textbf{0.9511}  \\
\hline
\end{tabular}
}
\caption{Quantitative comparison on Epic100, EgoFHO, DualArm, and Bridge, reporting the FVD, VTQ, VTC, and VIC scores.}
\label{tab:compare}
\vspace{-0.1in}

\end{table*}

\subsection{EIVST Datasets}
\label{sec:dataset}
We conduct experiments on two egocentric human–object interaction datasets and two robotic manipulation datasets: EpicKitchens-100 (Epic100)~\cite{epic}, Ego4D FHO (EgoFHO)~\cite{grauman2022ego4d}, DualArm~\cite{aist2025bimanip,agibot,wu2025robomind} and Bridge~\cite{ebert2021bridge}. 
\textbf{Epic100} captures daily human activities in the kitchen from a first-person perspective, providing fine-grained action annotations for each video clip. \textbf{EgoFHO} contains first-person daily-life activity videos across a variety of scenarios, including indoor and outdoor, with human-annotated descriptions. \textbf{DualArm} is collected from multiple bimanual manipulation datasets~\cite{aist2025bimanip,agibot,wu2025robomind} using a dual-arm robot. \textbf{Bridge} captures video clips from a single-arm robotic platform. From the original annotated videos, we extract sub-clips using a random frame stride. Each sub-clip is then paired with the original action instruction and re-captioned using Qwen2.5-VL to ensure alignment between the transition process and the action instruction. For each dataset, we use 90\% of the data for training and select 1,000 samples from the remaining unseen 10\% for evaluation.

\subsection{Implementation Details}
\label{sec:impl_detail}
\noindent \textbf{Training Details of TransitionVLM.} 
We generate about 10k state-aware descriptions and 30k transition-aware descriptions for VLM tuning using a mixed dataset composed of Epic100, EgoFHO. For the robotic datasets, \textit{i.e.,} DualArm and Bridge, we generate about 2k state-aware and 4k transition-aware descriptions for VLM tuning. These descriptions are obtained by querying GPT-4o with specifically designed inputs. The learning rate is set to $1 \times 10^{-5}$ for the Adapter module and $5 \times 10^{-5}$ for the LoRA module, respectively. We configure the LoRA module with $\text{lora\_rank} = 64$ and $\text{lora\_alpha} = 32$. The total number of training epochs is set to 4, with a batch size of 128.  

\begin{table}[t]
\centering

\setlength{\tabcolsep}{2pt}
\resizebox{0.45\textwidth}{!}{
\begin{tabular}{c|cccccc}
\hline
\multirow{2}{*}{Method} & \multicolumn{6}{c}{\textbf{Epic100}}\\ 
 &  FILM & TRF & GI  & SEINE  & DC-Interp & \textbf{Ours} \\
\hline
Reason. $\uparrow$ &$1.60\%$ &$1.20 \%$  &$4.13\%$ &$7.20\%$  & $10.95\%$  &$74.92\%$ \\
Align. $\uparrow$ &$1.73\%$  &$1.62\%$  & $4.12\%$             &$8.93 \%$ & $12.75\%$ &$70.85\%$  \\
Motion. $\uparrow$ & $0.92\%$  &$1.52\%$  &  $2.02\%$ & $6.85 \%$  & $11.63\%$ &$77.06\%$ \\ 
\hline
\end{tabular}
}
\caption{User preference evaluation in the reasonability of transition steps (Reason.), their alignment with the instruction (Align.), and motion quality (Motion.).}
\vspace{-0.3in}
\label{tab:user_study}
\end{table}

\noindent \textbf{Training Details of Video Generation Model.} The proposed EgoIn framework is built upon the DynamiCrafter Interpolation (DC-Interp)~\cite{xing2025dynamicrafter}, a lightweight and GPU memory-friendly method. We set the number of layers $M_1$ and $M_2$ used in Multi-modal Semantic State Encoding to 2. We adopt 16 queries in SQ-Former and set the feature lengths of $\widetilde{F}^{S}$ and $\{F_j^T\}_{j=1}^K$ to 32 and 77, respectively. During training, we first optimize the TC module for 20k steps with a learning rate of $1 \times 10^{-4}$ and a batch size of 64 while keeping all other parameters frozen, which helps prevent training instability. We then jointly fine-tune the spatial layers of the U-Net together with the TC module for an additional 10k steps using a learning rate of $2 \times 10^{-5}$ and a batch size of 32. During inference, we apply DDIM sampling with 50 steps.

\noindent \textbf{Evaluation Metrics.} To evaluate the video generation performance, we use the Fréchet Video Distance (FVD)~\cite{unterthiner2019fvd}, video temporal quality (VTQ), video-text consistency (VTC), and video-image consistency (VIC) scores, following the VBench~\cite{huang2024vbench++} protocol.

\subsection{Comparison with State-of-the-Arts.}
\label{sec:quant_res}

\begin{table}[t]
\centering

\setlength{\tabcolsep}{2pt}
\resizebox{0.5\textwidth}{!}{
\begin{tabular}{c|cccc|cccc}
\hline
\multirow{2}{*}{Method} & \multicolumn{4}{c|}{\textbf{Epic100}} & \multicolumn{4}{c}{\textbf{Bridge}} \\
 & FVD$\downarrow$ & VTQ$\uparrow$ & VTC$\uparrow$ & VIC$\uparrow$  & FVD$\downarrow$ & VTQ$\uparrow$ & VTC$\uparrow$ & VIC$\uparrow$ \\
\hline Baseline & 296.67 & 0.8797 & 0.2041 & 0.9081 &  287.47 & 0.9153 & 0.2092 & 0.9302 \\
\hline
+ TVLM &  261.78 & 0.8909 & 0.2233 & 0.9146 &  252.02 & 0.9235 & 0.2258 & 0.9348 \\
+ TVLM \& TC & 232.10 & 0.9013 & 0.2312 & 0.9251 &  220.92 &  0.9316 & 0.2337 & 0.9435 \\
+ TVLM \& TC \&  OAS & 215.27 & 0.9081 & 0.2373 & 0.9313 & 191.09 & 0.9374 & 0.2395 & 0.9511 \\ \hline
\end{tabular}
}
\caption{Ablation on each component of egoIn. ``TVLM'' and ``TC'' represents the TransitionVLM and Transition Conditioning module, respectively.}
\label{tab:ablation}
\end{table}

\begin{table}[!t]
\centering
\vspace{-0.1in}
\setlength{\tabcolsep}{2pt}
\resizebox{0.48\textwidth}{!}{
\begin{tabular}{c|cccc|cccc}
\hline
\multirow{2}{*}{\textbf{K}} & \multicolumn{4}{c|}{\textbf{Epic100}} &  \multicolumn{4}{c}{\textbf{EgoFHO}}\\ 
 &  FVD$\downarrow$ & VTQ$\uparrow$ & VTC$\uparrow$ & VIC$\uparrow$  &  FVD$\downarrow$ & VTQ$\uparrow$ & VTC$\uparrow$ & VIC$\uparrow$ \\
\hline
$K=1$ & 247.41 & 0.8934 & 0.2247 & 0.9256 & 241.53 & 0.8868 & 0.2228 & 0.9349 \\
$K=2$ & 228.66 & 0.8996 & 0.2319 & 0.9273 & 227.91 & 0.8935 & 0.2267 & 0.9318 \\
$K=4$ & 233.25 & 0.9013 & 0.2282 & 0.9238 & 230.90 & 0.8908 & 0.2284 & 0.9333 \\
$K\in[1,4]$ & 215.27 & 0.9081 & 0.2373 & 0.9313 & 203.85 & 0.8987 & 0.2340 & 0.9396 \\
\hline
\end{tabular}
}
\vspace{-1mm}
\caption{Ablation on transition steps.}
\vspace{-5mm}
\label{tab:k_steps_num}
\end{table}
We compare our method with various state-of-the-art methods. FILM~\cite{reda2022film}, TRF~\cite{feng2025explorative}, and GI~\cite{wang2024generative} are video interpolation methods without using textual guidance. SEINE~\cite{chen2023seine} is designed for video transition, while DC-Interp~\cite{xing2025dynamicrafter} is a text-guided frame interpolation model. For a fair comparison, we fine-tune the foundation models used by TRF and other methods using our EIVST training dataset.
The quantitative results are presented in Tab.~\ref{tab:compare}. Our method EgoIn significantly outperforms other I2V methods in FVD, VTQ, VTC, and VIC scores across four datasets, highlighting its capability in generating stable and smooth video transitions while better aligning with both textual and visual instructions. We also conducted a user study to evaluate the reasonability of the generated transition steps, their alignment with action instructions, and motion quality, as shown in Tab.~\ref{tab:user_study}. This study involved 40 participants comparing our method with other methods. Our method received over 50\% of the votes, outperforming other methods and making it the most preferred for generating visually appealing videos with reasonable transitions. The qualitative results are illustrated in Fig.~\ref{fig:compare}. We observe that FILM, GI and SEINE achieve visually smooth transitions. They struggle to generate actions in complex object state transitions. DC‑Interp tends to generate videos that are naive and straightforward, while missing the key transition steps required to reach the target state from the initial state. The proposed EgoIn generates more reasonable intermediate state transitions, while maintaining high temporal quality and strong video-condition consistency. 

\subsection{Ablation Study}
\label{sec:abl}

\begin{figure}[t]
	\centering
	\includegraphics[width=0.5\textwidth]{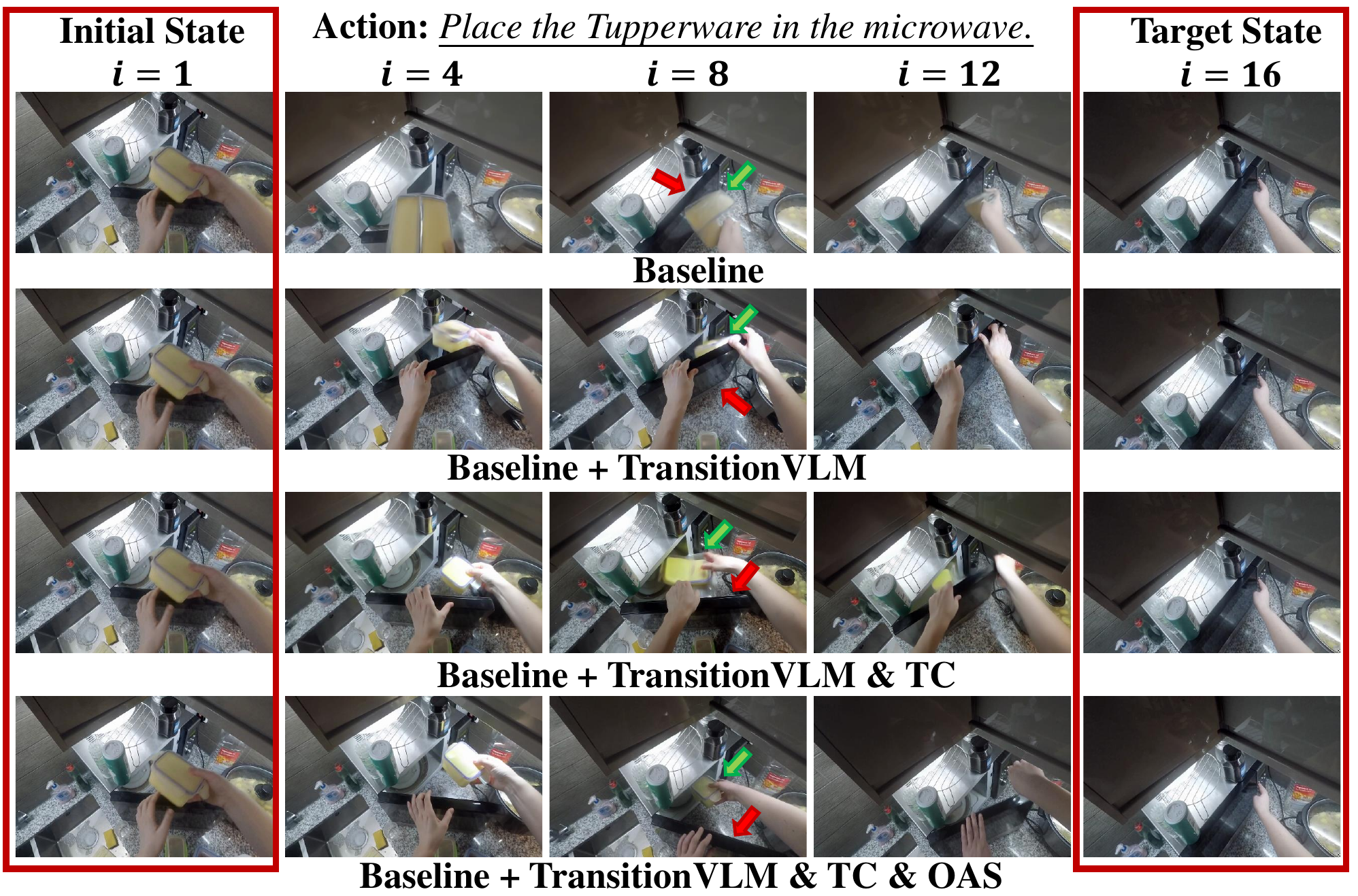}
	\caption{Ablation on the components of EgoIn.}
	\label{fig:ablation}
    \vspace{-0.25in}
\end{figure}

\noindent\textbf{Ablation on the components of EgoIn.}
The quantitative and qualitative results are presented in Tab.~\ref{tab:ablation} and Fig.~\ref{fig:ablation}, respectively. We use DC-Interp, fine-tuned on our dataset, as the baseline. When the baseline is equipped with TransitionVLM, the model receives extended textual conditioning by concatenating descriptions of the initial state, target state, and transition steps generated by TransitionVLM. The videos generated by this model follow more reasonable transition steps than the baseline (e.g., the hands first open the microwave door and then place the Tupperware inside). By further incorporating frame-wise transition conditions into the diffusion model through the TC module, the transition steps become more controllable and realistic, with more accurate temporal allocation of each step (e.g., the microwave door gradually closes after the Tupperware is placed inside). Additionally, applying the proposed OAS during training ensures consistent appearance and smooth motion of both the Tupperware and the microwave across frames, further improving the quality of the generated videos.

\noindent\textbf{Ablation on transition steps.}
We compare three fixed-step configurations for transition descriptions in Tab.~\ref{tab:k_steps_num}. In the $K{=}1$ setting, performance degrades due to the \textbf{limited context of single-step transitions}, which tends to be insufficient for guiding the video model in generating coherent transitions between initial and target frames. Increasing the number of steps to $K{=}2$ yields clear improvements, as the additional step provides richer context for describing the state transition. Although $K{=}4$ provides finer-grained temporal coverage, it tends to introduce redundant steps for simpler transitions, which may confuse the video generation model and degrade visual quality.

\begin{figure}[t]
	\centering
	\includegraphics[width=0.5\textwidth]{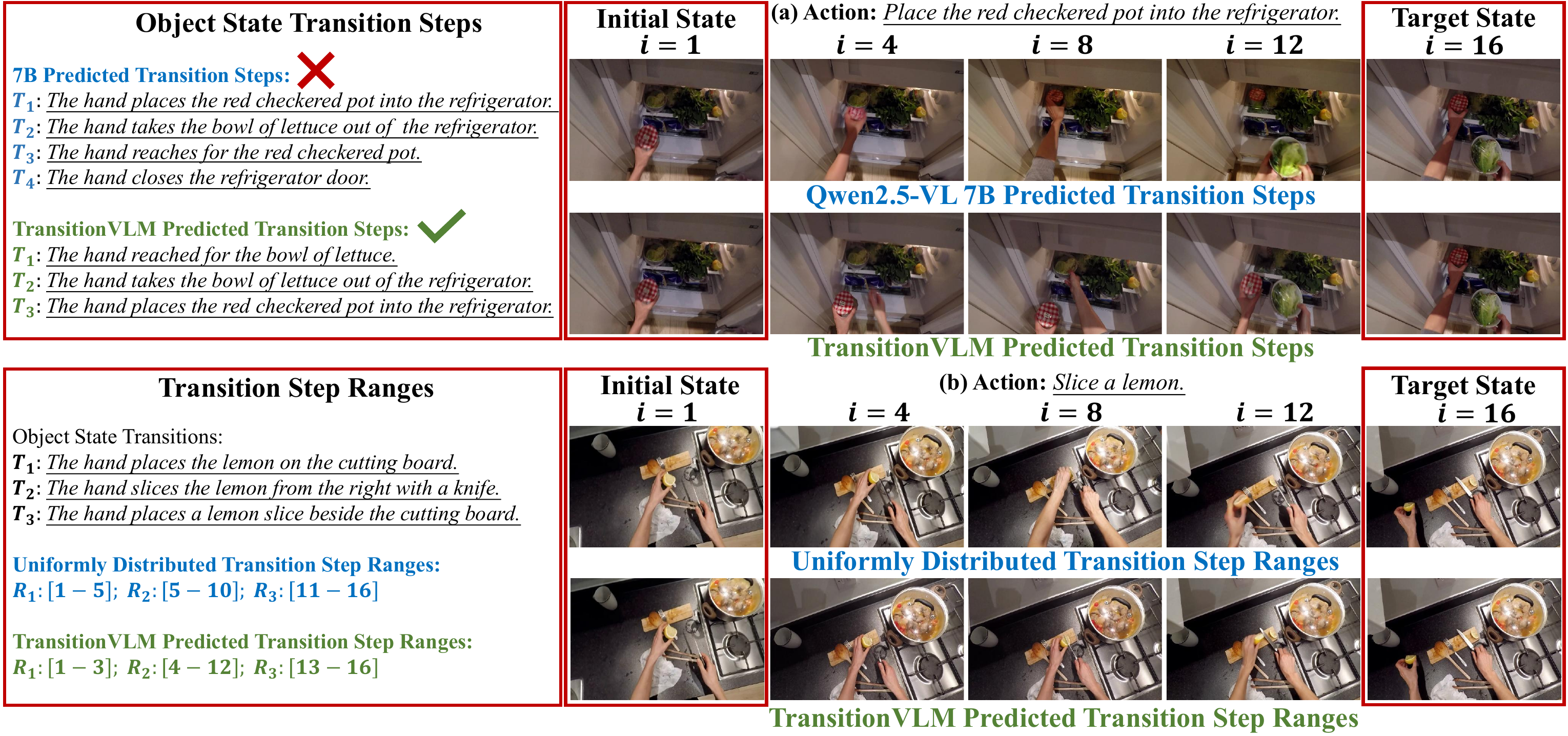}
        \vspace{-7mm}
	\caption{Ablation on the effectiveness of TransitionVLM in video generation.}
	\label{fig:VLM-case}
 		\vspace{-0.2in}
\end{figure}

\noindent{\textbf{Ablation on the effectiveness of TransitionVLM in video generation.}}  We conduct two toy experiments to evaluate the effectiveness of TransitionVLM. The first experiment fine-tunes the I2V model with transition steps generated by either the original VLM or TransitionVLM, which has been tuned on curated data. As shown in Fig.~\ref{fig:VLM-case} (a), video sequences conditioned on transitions from the original VLM show unrealistic state changes. For instance, in the first step, the red checkered pot is placed into a refrigerator already full, leading to unnatural object overlaps. Additionally, spurious details in the final step introduce ambiguity, degrading the model's ability to generate coherent videos. The second experiment examines how temporal range settings affect video generation. We compare two settings: (1) uniformly distributing temporal ranges based on the number of transition steps, and (2) predicting transition ranges using TransitionVLM. As shown in Fig.~\ref{fig:VLM-case} (b), videos generated with predicted temporal ranges exhibit more natural, temporally aligned, and semantically coherent transitions

\section{Conclusion}
Modeling object state transitions from an egocentric perspective enables machines to better mimic human understanding of physical transformation processes and reduce the embodiment gap. We define this modeling process as Egocentric Instructed Visual State Transition (EIVST). To achieve this, we propose the EgoIn framework, which integrates TransitionVLM for inferring transition steps from the action instruction and Transition Conditioning for generating frame-wise transition conditions. We further introduce Object-aware Auxiliary Supervision to enhance appearance consistency and ensure smooth motion during transitions. Extensive experiments on human–object and robot–object interaction datasets demonstrate that our method produces semantically meaningful and visually coherent transformation sequences, outperforming methods tailored for video prediction and video frame interpolation even when these methods are fine-tuned on the egocentric dataset.
\section{Limitation and Future Work}
Generating long-horizon state transitions that involve substantial scene or viewpoint changes remains a challenging problem for EgoIn. Addressing such complex transitions requires more robust modeling of long-range dependencies and stronger multi-view consistency. We consider this an important direction for future work and plan to further improve our framework to better handle large viewpoint shifts and dynamic environments.

{
\small
\bibliographystyle{ieeenat_fullname}
\bibliography{main}
}

\end{document}